\documentclass[journal]{IEEEtran}
\usepackage{amsmath,amsfonts}
\usepackage{algorithmic}
\usepackage{algorithm}
\usepackage{array}
\usepackage[caption=false,font=normalsize,labelfont=sf,textfont=sf]{subfig}
\usepackage{textcomp}
\usepackage{stfloats}
\usepackage{url}
\usepackage{verbatim}
\usepackage{graphicx}
\usepackage{cite}

\usepackage{multirow}
\usepackage{booktabs}
\usepackage{stfloats}

\usepackage{etoolbox}
\makeatletter
\patchcmd{\@makecaption}
  {\scshape}
  {}
  {}
  {}
\makeatother

\begin{document}

\title{PPC-MT: Parallel Point Cloud Completion with Mamba-Transformer Hybrid Architecture}

\author{\IEEEauthorblockN{Jie Li\IEEEauthorrefmark{1}, 
                     Shengwei Tian\IEEEauthorrefmark{1}, 
                     Long Yu\IEEEauthorrefmark{1}, and
                     Xin Ning\IEEEauthorrefmark{2}}
    \IEEEauthorblockA{\IEEEauthorrefmark{1}Xinjiang University, Urumqi, China}
    \IEEEauthorblockA{\IEEEauthorrefmark{2}Institute of Semiconductors, Chinese Academy of Sciences, Beijing, China}
}



\maketitle

\begin{abstract}
Existing point cloud completion methods struggle to balance high-quality reconstruction with computational efficiency. To address this, we propose PPC-MT, a novel parallel framework for point cloud completion leveraging a hybrid Mamba-Transformer architecture. Our approach introduces an innovative parallel completion strategy guided by Principal Component Analysis (PCA), which imposes a geometrically meaningful structure on unordered point clouds, transforming them into ordered sets and decomposing them into multiple subsets. These subsets are reconstructed in parallel using a multi-head reconstructor. This structured parallel synthesis paradigm significantly enhances the uniformity of point distribution and detail fidelity, while preserving computational efficiency. By integrating Mamba’s linear complexity for efficient feature extraction during encoding with the Transformer’s capability to model fine-grained multi-sequence relationships during decoding, PPC-MT effectively balances efficiency and reconstruction accuracy. Extensive quantitative and qualitative experiments on benchmark datasets, including PCN, ShapeNet-55/34, and KITTI, demonstrate that PPC-MT outperforms state-of-the-art methods across multiple metrics, validating the efficacy of our proposed framework. 
\end{abstract}


\section{Introduction}
Point clouds, as a core representation of 3D scenes and objects, play an indispensable role in fields like autonomous driving, robot navigation, augmented reality, and 3D modeling \cite{guo2020deep,chen20203d,pomerleau2015review,chen2019overview,li2022high}. However, due to limitations in sensor precision, occlusions between objects, and viewpoint constraints, point cloud data acquired in real-world scenarios often suffer from incompleteness or missing regions. These data imperfections significantly limit the performance of downstream tasks such as object detection, scene understanding, and pose estimation. Consequently, developing efficient and robust point cloud completion algorithms has become a critical research direction in 3D computer vision. 

Over the years, as research in point cloud completion has progressed, neural network methods based on encoder-decoder architectures have gradually become the mainstream approach. Early single-stage methods \cite{yang2017foldingnet,tchapmi2019topnet,liu2020morphing,groueix2018papier} typically extract global features from incomplete input point clouds through encoders, followed by decoders that directly map features to generate complete point clouds. However, due to their over-reliance on global features, single-stage methods have relatively limited capabilities in recovering fine local details and handling large-scale missing regions, with generalization performance often facing challenges. With PCN's pioneering introduction \cite{yuan2018pcn} of a two-stage point cloud completion strategy, features are first extracted from incomplete point clouds through an encoder to generate a low-resolution coarse point cloud as a shape skeleton, and then based on this foundation, combined with encoder features, a decoder further refines to generate high-resolution complete point clouds. Inspired by this approach, subsequent research \cite{wen2021pmp,wang2021cascaded,huang2020pf} has further explored multi-stage cascaded decoding or hierarchical generation strategies, improving completion accuracy and robustness through multi-step iteration or hierarchical refinement. Although multi-stage methods typically outperform single-stage and two-stage methods in terms of performance, when the intermediate point cloud representations rely on a large number of points, computational costs and complexity increase significantly, posing challenges for practical applications.

\begin{figure}
  \centering
  \includegraphics[width=\linewidth]{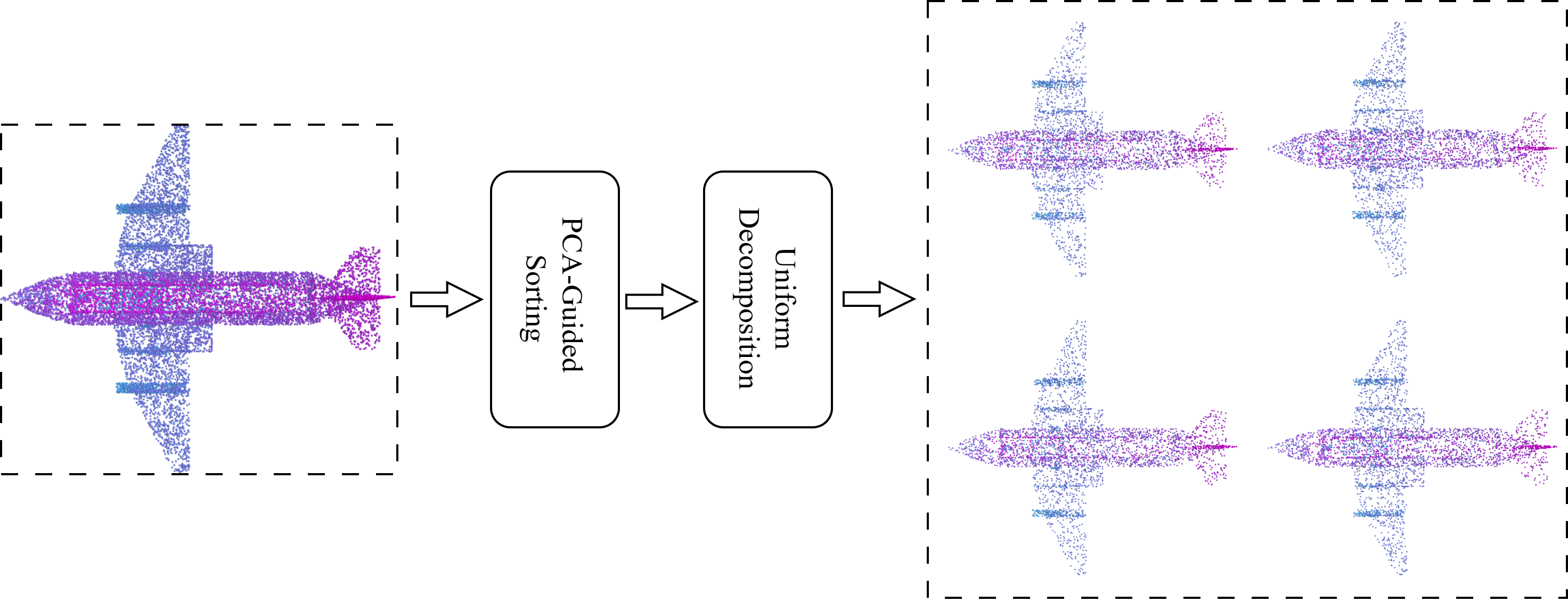}
  \caption{Illustration of PCA-Guided Decomposition. Target point clouds are first sorted according to Principal Component Analysis (PCA) and then uniformly decomposed into several structured segments.}
  \label{fig:decomposition}
\end{figure}

To overcome the bottlenecks of existing point cloud completion methods in handling complex missing regions and balancing quality with efficiency, this paper proposes the Parallel Point Cloud Completion framework PPC-MT. Inspired by the divide-and-conquer approach, we decompose the goal of reconstructing a complete point cloud into manageable sub-problems. As shown in Fig.~\ref{fig:decomposition}, we use Principal Component Analysis (PCA) to guide the uniform decomposition of a target point cloud into multiple sub-target point clouds for separate reconstruction, ultimately completing the reconstruction of the original point cloud collectively. Furthermore, we introduce the state space model Mamba to improve feature extraction efficiency, while maintaining the use of Transformer structures to model relationships between multiple sequences to ensure feature reconstruction accuracy.

Existing methods employ two sets of quantitative evaluation systems when assessing point cloud completion results. For supervised completion datasets (such as PCN and ShapeNet55/34), most studies only adopt Chamfer Distance (CD)~\cite{fan2017point} and F-Score~\cite{tatarchenko2019fscore} as evaluation metrics, while for unsupervised completion datasets (such as KITTI), Fidelity and MMD~\cite{yuan2018pcn} are used as evaluation metrics. These two evaluation systems are actually neither comprehensive nor entirely suitable. Therefore, for the former, we believe that DCD~\cite{wu2021density} and EMD~\cite{fan2017point} should be additionally introduced to evaluate completion results, while for the latter, we provide additional evaluation methods including Consistency and Uniformity for reference only.

Overall, our contributions can be summarized as follows:
\begin{itemize}
    \item We propose a PCA-guided decomposition method that assigns order to unordered point clouds, uniformly decomposing target point clouds into multiple sub-targets, and design a multi-head parallel reconstruction mechanism based on this to overcome the limitations of traditional serial methods.
    \item We are the first to combine Mamba with Transformer for point cloud completion, leveraging Mamba's linear complexity to efficiently encode global context while using Transformer to precisely model relationships between multiple sequences.
    \item For both supervised and unsupervised datasets, we provide more comprehensive metric evaluations, aiming to conduct more robust assessments of completion quality from multiple dimensions including shape, detail, and distribution.
    \item Extensive experiments on multiple public point cloud completion datasets including PCN, ShapeNet-55/34, and KITTI demonstrate that PPC-MT achieves significant performance improvements compared to existing state-of-the-art methods.
\end{itemize}

\begin{figure*}
  \centering
  \includegraphics[width=\linewidth]{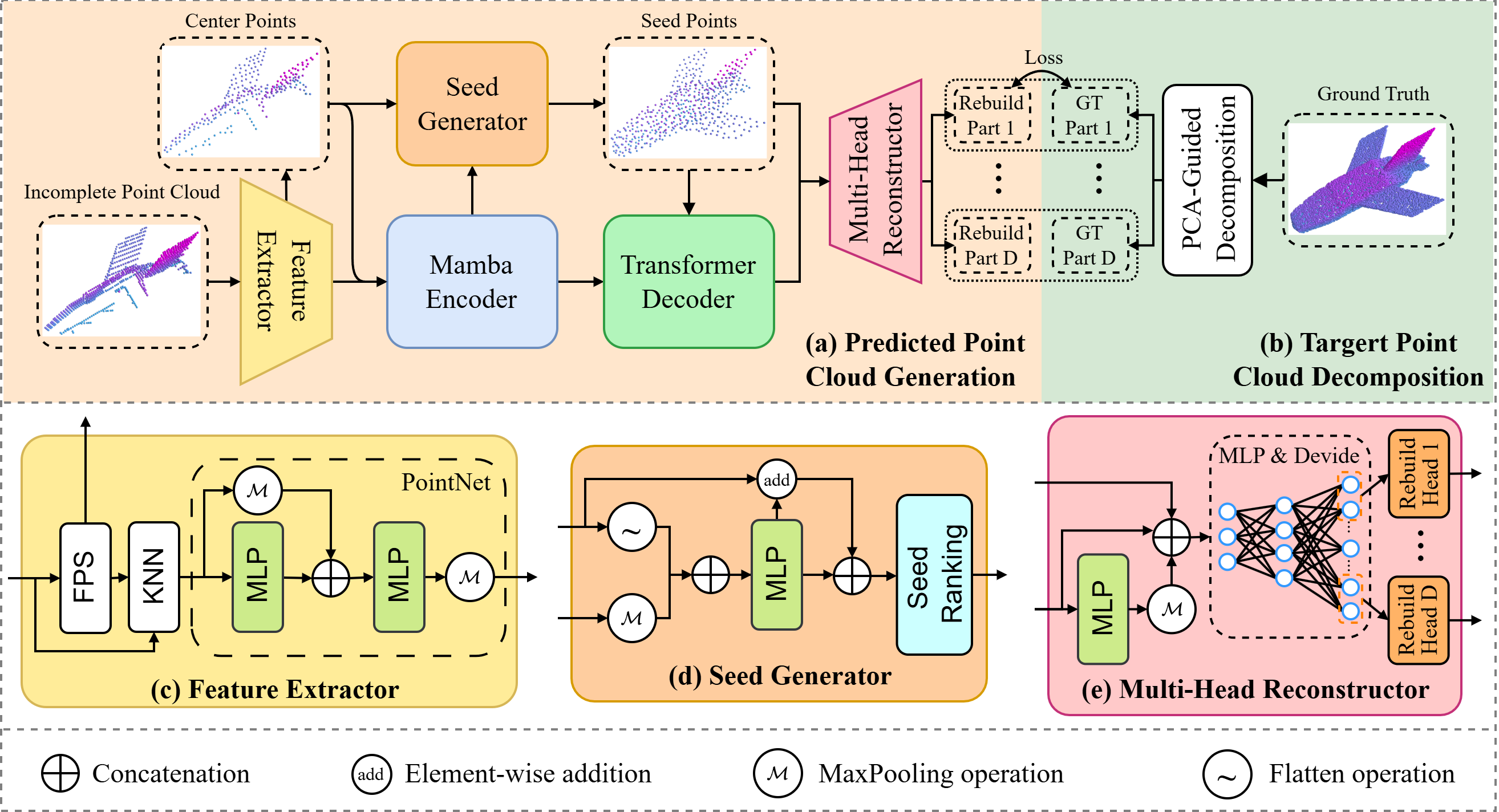}
  \caption{The PPCMT framework consists of two primary components: (a) Predicted Point Cloud Generation and (b) Target Point Cloud Decomposition. The Predicted Point Cloud Generation component includes multiple modules, three of which are illustrated in this figure: (c) Feature Extractor, (d) Seed Generator, and (e) Multi-Head Reconstructor.}
  \label{fig:ppcmt}
\end{figure*}

\section{Related Works}
\label{related}

\subsection{Point Cloud Completion}
Point cloud completion, a fundamental task in 3D vision, has garnered significant attention in recent years. Early works \cite{kazhdan2006poisson,wang2006fitting} primarily relied on geometric priors or traditional optimization methods, but these approaches showed limited performance when dealing with complex shapes and large-scale missing data. With the advancement of deep learning, neural network-based methods have become mainstream, broadly categorized into single-stage and multi-stage approaches. Single-stage completion methods typically employ an encoder-decoder architecture for direct end-to-end mapping from incomplete to complete point clouds. Representative works like FoldingNet \cite{yang2017foldingnet} and AtlasNet \cite{groueix2018papier} introduced deformable template ideas based on 2D parameterized manifolds, learning folding operations or mapping multiple surface patches to generate point clouds. TopNet \cite{tchapmi2019topnet} designed a tree-structured decoder with a hierarchical structure to generate point cloud structures. MSN \cite{liu2020morphing} generated multiple point cluster centers and sampled points to cover the complete shape. While single-stage methods are favored for their structural simplicity and high computational efficiency, the global feature vector can become an information bottleneck, limiting their ability to accurately reconstruct complex geometric details. To improve reconstruction quality, PCN \cite{yuan2018pcn} first proposed a "two-stage" strategy, generating a low-resolution coarse point cloud first, then refining it to a high-resolution result. This idea spurred a series of multi-stage methods. GRNet \cite{xie2020grnet} utilized voxel grids for progressive refinement. PoinTr \cite{yu2021pointr} introduced Transformer into the set-to-set translation process, capturing global-local correlations with self-attention. SnowflakeNet \cite{xiang2021snowflakenet} employed a "snowflake"-style point splitting mechanism to grow the point cloud progressively across layers. SeedFormer \cite{zhou2022seedformer} and AdaPoinTr \cite{yu2023adapointr} further enhanced detail preservation and shape consistency through seed point expansion and adaptive attention. ProxyFormer \cite{li2023proxyformer} optimized completion through proxy alignment and missing-part-sensitive Transformers, fusing existing and missing point features to generate high-precision point clouds while maintaining efficient inference. SVDFormer \cite{zhu2023svdformer} leveraged self-view enhancement and dual generators, combining multi-view global shape and local structural priors to achieve high-quality completion without requiring additional pairing information. The superior accuracy of these methods confirms the effectiveness of multi-stage approaches. The superior accuracy of these methods confirms the effectiveness of multi-stage approaches. However, starting from two-stage methods, many approaches increase the number of completion stages to improve accuracy. This serial approach significantly increases computational complexity. In contrast, the proposed PPC-MT aims to address this problem from a parallel perspective.

\subsection{State Space Models}
State Space Models (SSM) have emerged as a highly promising technology in sequence modeling, modeling complex systems by connecting inputs and outputs through hidden states \cite{xu2024visual}. Inspired by Structured State Space (S4) \cite{gu2021efficiently} models, several studies \cite{smith2022simplified,fu2022hungry} have explored using SSM to model long-range dependencies. The recently proposed Mamba model \cite{gu2023mamba}, in particular, has gained attention for its outstanding performance in global perception. Compared to Transformer's $O(n^2)$ computational complexity, Mamba achieves $O(n)$ complexity, demonstrating linear scalability when processing long sequences. Inspired by Mamba's success, researchers have proposed various methods to validate its effectiveness in 2D vision tasks \cite{zhu2024visionmamba,liu2024vmambavisualstatespace,zhou2024umamba}. In the 3D point cloud domain, effectively combining SSMs, originally designed for sequential data, with the inherently unordered nature of point clouds presents a significant challenge. PointMamba \cite{liang2024pointmamba} and PCMamba \cite{zhang2024pointcloudmamba} proposed different strategies to convert point clouds into ordered structures: PointMamba reorders points along coordinate axes, while PCMamba reorganizes data using various ordering methods like z-order and Hilbert order. Mamba3D \cite{han2024mamba3d} implemented bidirectional modeling through a channel flipping strategy, effectively addressing potential biases in Mamba caused by pseudo-sequential dependencies. Furthermore, PoinTramba \cite{wang2024pointramba} first validated the feasibility of combining Transformer and Mamba in the 3D point cloud domain. These methods have achieved notable results in point cloud classification and segmentation tasks. However, the application of Mamba in point cloud completion tasks remains largely unexplored. Specifically, how to effectively integrate the respective strengths of Transformer and Mamba to achieve superior performance and lower computational complexity is still an open question requiring investigation.

\section{Methods}

\subsection{Overview}
PPC-MT consists of two primary components: Predicted Point Cloud Generation (Fig.~\ref{fig:ppcmt} (a)) and Target Point Cloud Decomposition (Fig.~\ref{fig:ppcmt} (b)). The prediction part includes five modules: Feature Extractor, Mamba Encoder, Seed Generator, Transformer Decoder, and Multi-Head Reconstructor. The decomposition part is based on the PCA method.
Similar to SeedFormer and AdaPoinTr, we formulate the point cloud completion task as a set-to-set translation problem. Here, we refer to these sets as "point proxies", each representing a local region of the point cloud. Our goal is to transform the point proxies of the input point cloud into the point proxies of the predicted point cloud. Unlike previous methods, we use multiple heads to jointly predict the target point cloud.

Specifically, given an input point cloud $\mathcal{P}_{in} \in \mathbb{R}^{N\times3}$, the Feature Extractor first partitions it into G local groups using Farthest Point Sampling (FPS) and K-Nearest Neighbors (KNN) and extracts initial point proxy features $\mathcal{F}=\{F_1,F_2,\ldots,F_G\}$. Subsequently, we employ an efficient Mamba Encoder instead of a traditional Transformer to capture long-range dependencies within the point cloud, obtaining feature-enhanced point proxies $\mathcal{E}=\{E_1,E_2,\ldots,E_G\}$. Based on $\mathcal{E}$ and seed points $\mathcal{P}_0$ from Seed Generator, we generate $I$ seed point proxies $\mathcal{S}=\{S_1,S_2,\ldots,S_I\}$ which are then transformed into feature-enhanced point proxies $\mathcal{D}=\{D_1,D_2,\ldots,D_I\}$ via the Transformer Decoder. Finally, the Multi-Head Reconstructor reconstructs $\mathcal{D}$ into $U$ sets of predicted point proxies $\mathcal{O}=\{\mathcal{O}^{i}\}_{i=1}^U$, which jointly reconstruct the complete point cloud $\mathcal{P}=\bigcup_{i=1}^{U}{\mathcal{P}_{i}}$. To provide finer-grained supervision signals, we adopt a PCA-guided target point cloud decomposition strategy, breaking down the ground truth point cloud $\mathcal{G}$ into $U$ sub-point clouds $\{{\mathcal{G}_i}\}_{i=1}^{U}$. The entire framework is trained end-to-end using a multi-level loss function, including losses for center points, local point clouds, and the global point cloud.

\subsection{Predicted Point Cloud Generation}
\textbf{Feature Extractor.} The initial step reorganizes the input point cloud $\mathcal{P}_{in} \in \mathbb{R}^{N\times3}$ into point proxies to reduce subsequent computational complexity, a common operation in point cloud analysis \cite{pang2022masked,liang2024pointmamba,han2024mamba3d}. As detailed in Fig.~\ref{fig:ppcmt} (c), we first downsample $G$ center points $C$ from $\mathcal{P}_{in}$ using Farthest Point Sampling (FPS). Then, for each center point, we find its $K$ nearest neighbors using KNN to form $G$ point groups $P$. Points within each group are normalized relative to their corresponding center point. Finally, a lightweight PointNet[@PointNet] is used to generate the initial point proxies $\mathcal{F}=\{F_1,F_2,\ldots,F_G\}$:

\begin{equation}
    C=\mathrm{FPS}\left(\mathcal{P}_{in}\right),\qquad C\in \mathbb{R}^{G\times3}
\end{equation}
\begin{equation}
    P=\mathrm{KNN}\left(\mathcal{P}_{in},C\right),\qquad P\in \mathbb{R}^{G \times K \times3} 
\end{equation}
\begin{equation}
    \mathcal{F}=\mathrm{PointNet}(P),\qquad \mathcal{F}\in \mathbb{R}^{G \times D}
\end{equation}
where $D$ is the hidden dimension of the network. The lightweight PointNet primarily consists of MLPs and max-pooling operations.

\begin{figure}
  \centering
  \includegraphics[width=0.8\linewidth]{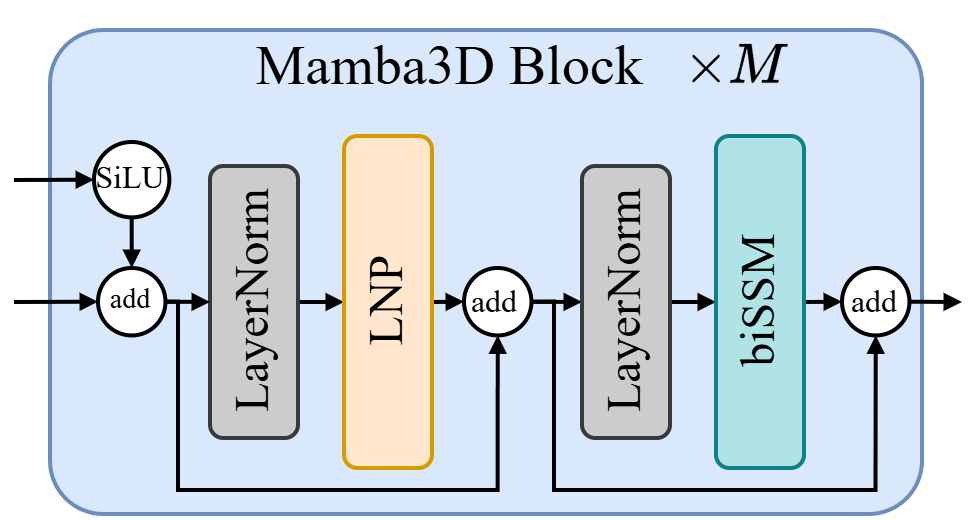}
  \caption{The detailed structure of the Mamba Encoder. }
  \label{fig:mamba3d}
\end{figure}

\textbf{Mamba Encoder.} Mamba \cite{gu2023mamba} was proposed to address the computational efficiency issues in modeling long sequences, achieving linear complexity $O(n)$. Mamba2 \cite{dao2024mamba2} further demonstrated its potential as an efficient alternative to self-attention. Considering that point proxies inherit the unordered nature of point clouds yet possess spatial coordinate correlations, Mamba3D \cite{han2024mamba3d} introduced Bidirectional-SSM and LNPBlock to specifically address potential pseudo-sequential dependencies and aggregate local spatial geometric features, proving their effectiveness experimentally. Therefore, we adopt Mamba3D Blocks to capture intra-group dependencies within the initial point proxies $\mathcal{F}$ and generate enhanced point proxy features $\mathcal{E}=\{E_1,E_2,\ldots,E_G\}$. Specifically, our Mamba Encoder, structured as shown in Fig.~\ref{fig:mamba3d}, contains M Mamba3D Blocks. Each Mamba3D Block includes two key components: LNP and biSSM. The pipeline can be represented by the following equations:
\begin{equation}
    z_0=\mathcal{F}+\mathrm{SiLU}(C),\qquad z_0\in \mathbb{R}^{G\times D}
\end{equation}
\begin{equation}
    {z^\prime}_l=\mathrm{LNP}(\mathrm{LN}(z_{l-1}+\mathrm{SiLU}(C)))+z_{l-1}, \qquad      l=1,\ldots
\end{equation}
\begin{equation}
    z_l=\text{bi-SSM}(\text{LN}(z^\prime_l))+z^\prime_l,\qquad l=1,\ldots,M
\end{equation}
\begin{equation}
    \mathcal{E}=z_M,\qquad \mathcal{E}\in \mathbb{R}^{G\times D}
\end{equation}

where $z_l$ is the output of layer $l$, and $\mathrm{LN}$ denotes LayerNorm. Notably, since $\mathcal{F}$ only represents local semantic patterns, each Mamba3D Block includes a standard learnable positional encoding $\mathrm{SiLU}(C)$ to explicitly encode the global positions of the point proxies, where the SiLU operation originates from \cite{hendrycks2016gaussian}.

\textbf{Seed Generator.} The Seed Generator aims to produce initial point cloud seeds, serving as the starting point for the decoder. These seed points are distributed at key locations of the target shape, laying a foundation for subsequent fine-grained generation. To fully leverage the coordinate information of the input point cloud, our proposed seed generation module adopts the design shown in  Fig.~\ref{fig:ppcmt} (d). Specifically, the module first applies max-pooling and flattening operations to the point proxy features $\mathcal{E}$ and center points $C$, respectively. Then, a MLP predicts an excessive number of candidate points ${\hat{\mathcal{P}}}_0\in R^{I^\prime\times3}$ (typically $I^\prime=768$). To enhance feature representation, residual connections are introduced for the top 256 predicted points. Next, a scoring network evaluates the quality of all candidate seed points, and the optimal set of seed points $\mathcal{P}_0\in R^{I\times3}$ (typically $I=512$) is selected based on the scores. This process is mathematically described as:

\begin{equation}
    {\hat{\mathcal{P}}}_0=\mathrm{MLP}\left(\mathcal{M}\left(\mathcal{E}\right), \mathrm{Flatten}\left(C\right)\right), \qquad {\hat{\mathcal{P}}}_0\in \mathbb{R}^{I^\prime\times3}
\end{equation}

\begin{equation}
    \mathcal{P}_0=\mathrm{TopK}(\mathcal{P}_0,\Omega(\mathcal{P}_0)), \qquad P_0 \in \mathbb{R}^{I\times3}
\end{equation}
where $\mathcal{M}$ is the max-pooling operation, $\mathrm{Flatten}$ is the flattening operation, $\Omega$ represents a scoring network composed of $\mathrm{MLP}$ and $\mathrm{Sigmoid}$ operations, and $\mathrm{TopK}$ denotes the operation for selecting points with the highest scores.

\begin{figure}
  \centering
  \includegraphics[width=0.5\linewidth]{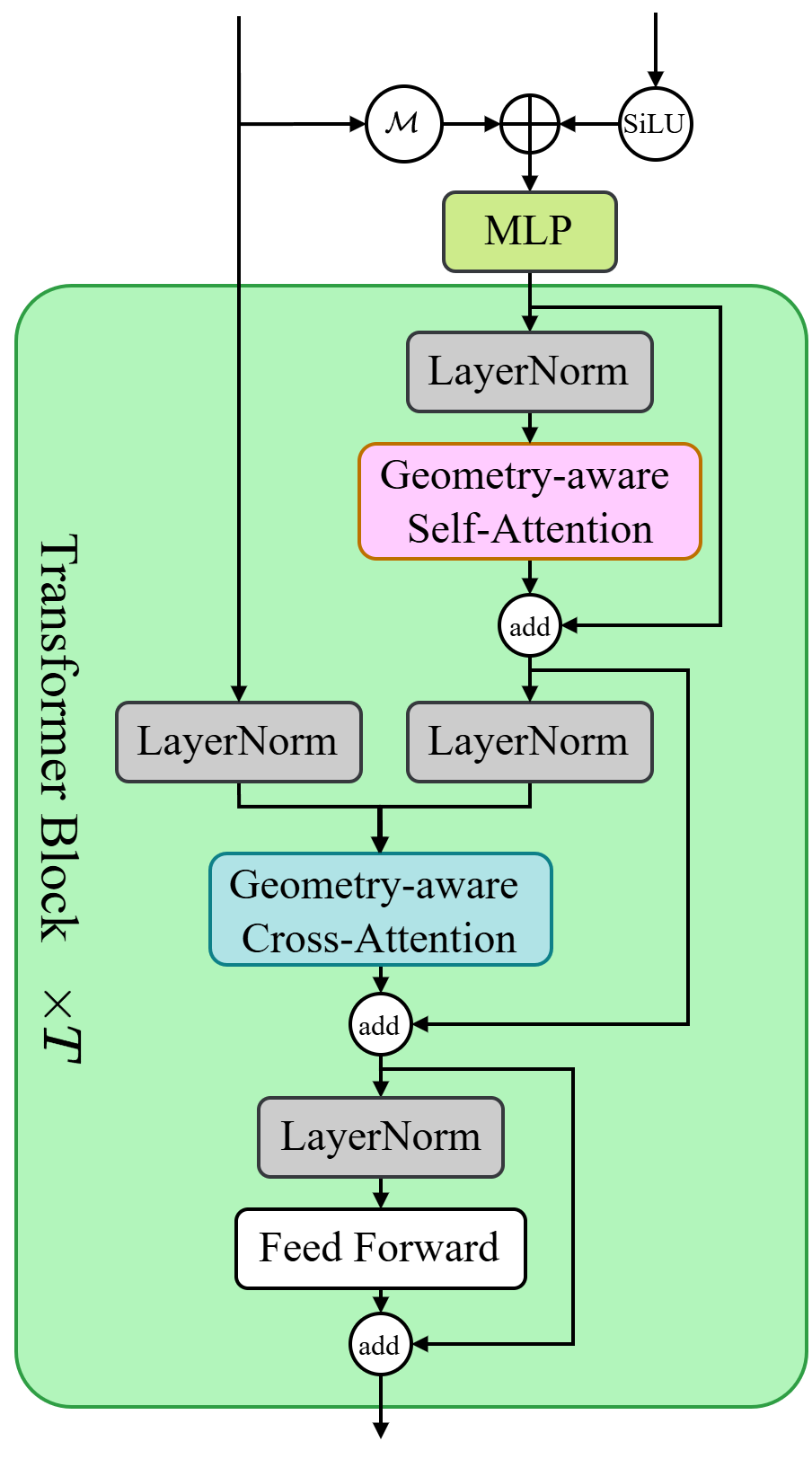}
  \caption{The detailed structure of the Transformer Decoder.}
  \label{fig:transformer}
\end{figure}

\textbf{Transformer Decoder.} Our module borrows from AdaPoinTr \cite{yu2023adapointr}, incorporating geometry-aware attention to adapt to the unordered nature of point cloud data. We continue to use Transformer in the decoder because we need the cross-attention mechanism to explicitly model the relationships between the point proxy features $\mathcal{E}=\{E_1,E_2,\ldots,E_G\}$ from the Mamba3D Encoder and the seed point proxies $\mathcal{S}=\{S_1,S_2,\ldots,S_I\}$. As shown in Fig.~\ref{fig:transformer}, our Transformer Decoder progressively transforms the seed point proxies $\mathcal{S}$ into point proxies $\mathcal{D}=\{D_1,D_2,\ldots,D_I\}$ through stacked attention layers. Each layer contains geometry-aware self-attention and cross-attention modules. The former allows seed point proxies $\mathcal{S}$ to exchange information among themselves, while the latter aims to extract relevant features from the encoder's point proxy features $\mathcal{E}$. The process is formulated as:

\begin{equation}
    w_0=\mathcal{S}=\mathrm{MLP}(\mathcal{M(E)},\mathrm{SiLU}(P_0)), \qquad w_0 \in \mathbb{R}^{I\times D}
\end{equation}
\begin{equation}
    w_l^\prime=w_{l-1}+ \mathrm{SelfAtt}(\mathrm{LN}(w_{l-1})),\qquad l=1,\ldots,T
\end{equation}
\begin{equation}
    w_l^{\prime\prime}=w_l^\prime+\mathrm{CrossAtt}(\mathrm{LN}(w_l^\prime), \mathrm{LN}(\mathcal{E})),\qquad l=1,\ldots,T
\end{equation}
\begin{equation}
    w_l=w_l^{\prime\prime}+\mathrm{FFN}(\mathrm{LN}(w_l^{\prime\prime})), \qquad l=1,\ldots,
\end{equation}
\begin{equation}
    \mathcal{D}=w_T, \qquad \mathcal{D}\in \mathbb{R}^{I\times D}
\end{equation}
where $\mathrm{SelfAtt}$ and $\mathrm{CrossAtt}$ represent geometry-aware self-attention and cross-attention, respectively. For more details on the geometry-aware attention module, please refer to AdaPoinTr.

\textbf{Multi-Head Reconstructor.} To correspond with the decomposed target point clouds described in Section~\ref{para:target}, we propose a Multi-Head Reconstructor, illustrated in  Fig.~\ref{fig:ppcmt} (e). Its core component is the MLP \& Divide module, which reshapes the Transformer's point proxy features $\mathcal{D}$ into $U$ sets of point proxies $\mathcal{O}=\{\mathcal{O}^{i}\}_{i=1}^U$. This allows these point proxies to be mutually independent while maintaining their interrelations. Each reconstruction head corresponding to a point proxy set $\mathcal{O}^{i}$ learns local coordinate offsets relative to the seed points $\mathcal{P}_0$. Adding these offsets to the seed points yields $U$ predicted point clouds $\{\mathcal{P}_i\}_{i=1}^U$. Merging all predicted point clouds gives the final complete predicted point cloud $\mathcal{P}_{out}\in R^{N_c\times 3}$. The advantage of this multi-head design is that it distributes the difficulty of learning global features across multiple reconstruction heads, thereby improving reconstruction accuracy and completeness. The process can be represented by the following formulas:
\begin{equation}
    \mathcal{O}=\psi([\mathcal{M}(\mathrm{MLP}(D)),\mathcal{D},P_0]), \qquad \mathcal{O} \in \mathbb{R}^{U\times I\times D}
\end{equation}
\begin{equation}
    \mathcal{P}_i=\mathcal{P}_0+\phi_i(\mathcal{O}^i),\qquad \mathcal{P}_i \in \mathbb{R} ^{\frac{N_c}{U}\times3} ,i=1,\ldots,U
\end{equation}
\begin{equation}
    \mathcal{P}=\bigcup_{i=1}^{U}{\mathcal{P}_{i}} , \qquad \mathcal{P} \in \mathbb{R}^{N_c\times3}
\end{equation}
where $\psi$ is the MLP \& Divide module, and $\phi_i$ is the $i$-th reconstruction head. Here, we employ simple MLP-based reconstruction heads. $N_c$ is the number of points in the complete output cloud. 

\subsection{Target Point Cloud Decomposition}
\label{para:target}
Existing point cloud completion methods typically rely on global supervision signals to guide the generation of the final point cloud. Such signals are often based on the average distance between all point pairs, like Chamfer Distance (CD) or Earth Mover's Distance (EMD). However, a single global supervision signal can easily lead the model into local optima, thereby limiting completion quality. Inspired by the divide-and-conquer strategy, as shown in Fig.~\ref{alg:pca_decomposition}, we propose a new method that decomposes the point cloud and applies supervision independently to each subset. 

The unordered nature of point clouds makes direct decomposition result in random and semantically meaningless subsets. To address this, we initially designed a point cloud sorting algorithm based on Principal Component Analysis (PCA). This algorithm first computes the three principal components of the target point cloud $\mathcal{G}$. Then, points are sorted according to the following rules: primarily based on their projection values onto the first principal component; if the first principal component projection values are identical, sorting is based on the second principal component projection values; if those are also identical, sorting is based on the third principal component projection values. This method transforms the unordered point cloud into an ordered point set, providing a structured basis for subsequent decomposition.

To better supervise point cloud generation, after obtaining the ordered point cloud, we further propose a uniform point cloud decomposition strategy. Since our point cloud completion task primarily relies on predicting offsets from seed points, we uniformly divide the sorted point cloud $\mathcal{G}_s$ into $U$ sub-point clouds, corresponding to the number of reconstruction heads. Specifically, each sub-point cloud $\mathcal{G}_i$ is generated by sampling points from the sorted point cloud $\mathcal{G}_s$ at intervals of $U$, ensuring the balance and representativeness of the subsets. Each sub-point cloud can receive independent supervision, thereby enhancing the precision of local geometric structures and the overall completion effect. The detailed steps for point cloud sorting and decomposition are provided in Algorithm~\ref{alg:pca_decomposition}.

\begin{algorithm}[t]
\caption{PCA-Guided Point Cloud Decomposition}
\label{alg:pca_decomposition}
\begin{algorithmic}[1]
    \REQUIRE Point cloud $\mathbf{\mathcal{G}} \in \mathbb{R}^{N_c \times 3}$, subset count $U$
    \ENSURE Subsets $\{\mathbf{\mathcal{G}}_1, \dots, \mathbf{\mathcal{G}}_U\}$
    
    \STATE $\boldsymbol{\mu} \gets \frac{1}{N_c} \sum_{i=1}^{N_c} \mathbf{\mathcal{G}}[i,:]$ \COMMENT{Centroid}
    \STATE $\mathbf{\mathcal{G}}_c \gets \mathbf{\mathcal{G}} - \boldsymbol{\mu}$ \COMMENT{Centralize}
    
    \STATE $\boldsymbol{\Sigma} \gets \frac{1}{N_c - 1} \mathbf{\mathcal{G}}_c^\top \mathbf{\mathcal{G}}_c$ \COMMENT{Covariance}
    \STATE $(\mathbf{V}, \boldsymbol{\Lambda}) \gets \mathrm{eig}(\boldsymbol{\Sigma})$ \COMMENT{Eigen decomposition}
    \STATE Sort $\mathbf{V}$ by descending $\boldsymbol{\Lambda}$
    
    \STATE $\mathbf{q} \gets \mathbf{V}^\top \mathbf{\mathcal{G}}_c$ \COMMENT{Project points}
    \STATE $\mathcal{B} \gets [(q_i, i) \mid q_i = \mathbf{q}[i,:], i=1,\dots,N_c]$
    \STATE Sort $\mathcal{B}$ by $q_i$ lexicographically
    \STATE $\mathcal{I} \gets [i_1, i_2, \dots, i_{N_c}]$ from $\mathcal{B}$
    
    \STATE $\mathbf{\mathcal{G}}_s \gets \mathbf{\mathcal{G}}[\mathcal{I},:]$ \COMMENT{Reorder}
    \FOR{$u = 0$ to $U-1$}
        \STATE $\mathbf{\mathcal{G}}_{u+1} \gets \{\mathbf{\mathcal{G}}_s[j,:] \mid j \bmod U = u, 1 \leq j \leq N_c\}$
    \ENDFOR
    
    \RETURN $\{\mathbf{\mathcal{G}}_1, \dots, \mathbf{\mathcal{G}}_U\}$
\end{algorithmic}
\end{algorithm}

\subsection{Flexible Training Loss}
Chamfer Distance (CD) is the most commonly used loss function in point cloud completion, known for its efficiency and invariance to order. It calculates the sum of squared nearest neighbor distances bidirectionally between the generated point cloud $\mathcal{P}$ and the ground truth point cloud $\mathcal{G}$:
\begin{equation}
   {CD}(\mathcal{P},\mathcal{G})=\frac{1}{\lvert\mathcal{P}\rvert}\sum_{p\in\mathcal{P}}\min_{g\in\mathcal{G}}{\lVert{p-g}\rVert_2}+\frac{1}{\lvert\mathcal{G}\rvert}\sum_{g\in\mathcal{G}}\min_{p\in\mathcal{P}}\lVert{g-p}\rVert_2 
\end{equation}
However, based on our prior work \cite{li2025flexible}, we found that the two components of CD have different focuses when guiding point cloud generation. We define:
\begin{equation}
    CD_{g}(\mathcal{P},\mathcal{G}) = \frac{1}{\lvert\mathcal{G}\rvert}\sum_{g\in\mathcal{G}}\min_{p\in\mathcal{P}}\lVert{g-p}\rVert_2
\end{equation}
\begin{equation}
    CD_{l}(\mathcal{P},\mathcal{G}) = \frac{1}{\lvert\mathcal{P}\rvert}\sum_{p\in\mathcal{P}}\min_{g\in\mathcal{G}}{\lVert{p-g}\rVert_2}
\end{equation}
Here, $CD_{g}$ calculates the distance from the ground truth points to the generated points, guiding the generation towards a good global distribution but prone to local optima. $CD_{l}$ calculates the distance from generated points to ground truth points and must be used in conjunction with $CD_{g}$ to guide the generation of superior local details. We believe that training with equal weights is not optimal; the use of these two components should be flexible and adaptable to the training objective, with $CD_{g}$ generally receiving higher weight. This paper does not primarily focus on discussing how to set the weights, so we simply apply generally suitable weight settings. 
In this paper, we have three targets requiring supervision: the seed points $\mathcal{P}_0$, the predicted points for each part $\{\mathcal{P}_i\}_{i=1}^U$, and the final predicted complete point cloud $\mathcal{P}$. Our requirement for the seed points $\mathcal{P}_0$ is only an initial global structure, so we use only $CD_{g}$. For the latter two targets, both global distribution and local structure are important. Therefore, our overall loss function is set as follows:
\begin{equation}
    \mathcal{L} = \mathcal{L}_{\mathcal{P}_0}+\frac{1}{U}\sum_{i=1}^{U}{\mathcal{L}_{\mathcal{P}_i}} + \mathcal{L}_{\mathcal{P}_{out}}
\end{equation}
where
\begin{equation}
    \mathcal{L}_{\mathcal{P}_0}=CD_g(\mathcal{P}_0,\mathcal{G})
\end{equation}
\begin{equation}
    {\mathcal{L}_{\mathcal{P}_i}}=CD_l(\mathcal{P}_i,\mathcal{G}_i) + 2CD_g(\mathcal{P}_i,\mathcal{G}_i)
\end{equation}
\begin{equation}
    {\mathcal{L}_{\mathcal{P}_{out}}}=CD_l(\mathcal{P}_{out},\mathcal{G}) + 2CD_g(\mathcal{P}_{out},\mathcal{G})
\end{equation}

\section{Evaluation}
In this section, we evaluate the completion capability of PPC-MT on two common point cloud completion benchmarks, PCN \cite{yuan2018pcn} and KITTI \cite{geiger2013kitti}. We also conduct training and testing on two additional datasets proposed by PoinTr \cite{yu2021pointr}, ShapeNet-55 and ShapeNet-34. Finally, through ablation studies, we demonstrate the effectiveness of the methods proposed in PPC-MT.

\subsection{Experiments on PCN Dataset}
\label{para-pcn}

\textbf{Dataset.} The PCN dataset, derived as an 8-category subset of the ShapeNet dataset, is widely used. Following previous work \cite{yuan2018pcn}, our training set contains 28,974 samples, and the test set includes 1,200 samples. Complete point clouds (16,384 points) are obtained by uniformly sampling from mesh model surfaces. Incomplete point clouds are generated by back-projecting depth maps from 8 different viewpoints and then padding to 2,048 points as input. 

\textbf{Evaluation Metrics.} Current works mostly focus on CD~\cite{fan2017point} and F-Score~\cite{tatarchenko2019fscore} metrics, which, however, do not fully reflect the quality of point cloud generation results. To obtain a clear and comprehensive comparison, we employ a wider range of complementary evaluation metrics. Specifically, we use Chamfer Distance \(\ell_1\) (CD-\(\ell_1\)), Earth Mover's Distance (EMD)~\cite{fan2017point}, Density-aware Chamfer Distance (DCD)~\cite{wu2021density}, and F-Score for quantitative evaluation. CD-\(\ell_1\) measures the average nearest neighbor distance between two point clouds, primarily focusing on the reconstruction quality of local geometric details; lower values indicate more accurate local reconstruction. EMD measures the minimum "cost" required to transform one point cloud into another, better reflecting the similarity of global shape distribution; lower is better. DCD is a density-aware variant of Chamfer Distance that considers the local density distribution, evaluating local details while ensuring regional density consistency, particularly suitable for evaluating non-uniformly sampled point clouds; lower is better. F-Score combines precision and recall, calculating the F1 score of point cloud matching given a threshold, balancing the evaluation of reconstruction completeness and accuracy; higher values generally indicate better overall reconstruction quality. Using these metrics in combination allows for a comprehensive evaluation of point cloud completion models from different perspectives.

\textbf{Quantitative comparison.} As shown in Table~\ref{tab:results-pcn}, PPC-MT achieves significant performance improvements on the PCN dataset. Specifically, our method attains state-of-the-art results on three key metrics: DCD (0.491), EMD (17.43), and F-Score (0.860), indicating that the point clouds generated by PPC-MT are geometrically closer to the ground truth. Although the CD-\(\ell_1\) score (6.60) is slightly higher than SVDFormer (6.54) and AdaPoinTr (6.53), the difference is minimal. Notably, compared to the current state-of-the-art method AdaPoinTr, our method improves F-Score by 1.5\% (from 0.845 to 0.860), reduces EMD by 27.7\% (from 24.12 to 17.43), and lowers DCD by 8.4\% (from 0.536 to 0.491). These results fully demonstrate the superiority of our proposed method in capturing both local details and global structure of point clouds, offering a new solution for the point cloud completion task. 

\begin{table}[]
\caption{Results on PCN dataset. (CD-\(\ell_1 \times 10^3\), EMD\(\times 10^3\) and F1-Score@1\%)}
\resizebox{\linewidth}{!}{%
\begin{tabular}{l|cccc}
\hline
\textbf{Method} & \textbf{CD-\(\ell_1\)\(\downarrow\)} & \textbf{DCD\(\downarrow\)} & \textbf{EMD\(\downarrow\)} & \textbf{F1\(\uparrow\)} \\ \hline
FoldingNet~\cite{yang2017foldingnet}      & 14.31           & -             & -             & -                 \\
TopNet~\cite{tchapmi2019topnet}          & 12.15           & -             & -             & -                 \\
AtlasNet~\cite{groueix2018papier}        & 10.85           & -             & -             & 0.616             \\
CRN~\cite{wang2021cascaded}            & 8.51            & -             & -             & 0.652             \\
PCN~\cite{yuan2018pcn}             & 9.64            & -             & -             & 0.695             \\
GRNet~\cite{xie2020grnet}           & 8.83            & 0.622         & -             & 0.708             \\
NSFA~\cite{zhang2021nsfa}            & 8.06            & -             & -             & 0.734             \\
PMP-Net++~\cite{wen2022pmp}       & 7.56            & 0.611         & -             & 0.781             \\
PoinTr~\cite{yu2021pointr}          & 7.26            & 0.571         & 24.57         & 0.797             \\
SnowflakeNet~\cite{xiang2021snowflakenet}    & 7.21            & 0.585         & -             & 0.801             \\
SeedFormer~\cite{zhou2022seedformer}      & 6.74            & 0.567         & 27.91         & 0.821             \\
ProxyFormer~\cite{li2023proxyformer}     & 6.77            & 0.555         & -             & -                 \\
SVDFormer~\cite{zhu2023svdformer}       & 6.54            & 0.532         & 24.58         & 0.841             \\
AdaPoinTr~\cite{yu2023adapointr}       & \textbf{6.53}            & 0.536         & 24.12         & 0.845             \\ \hline
PPC-MT (Ours)   & 6.60          & \textbf{0.491}   & \textbf{17.43}  & \textbf{0.860}         \\ \hline
\end{tabular}}
\label{tab:results-pcn}
\end{table}

\begin{figure*}
  \centering
  \includegraphics[width=\linewidth]{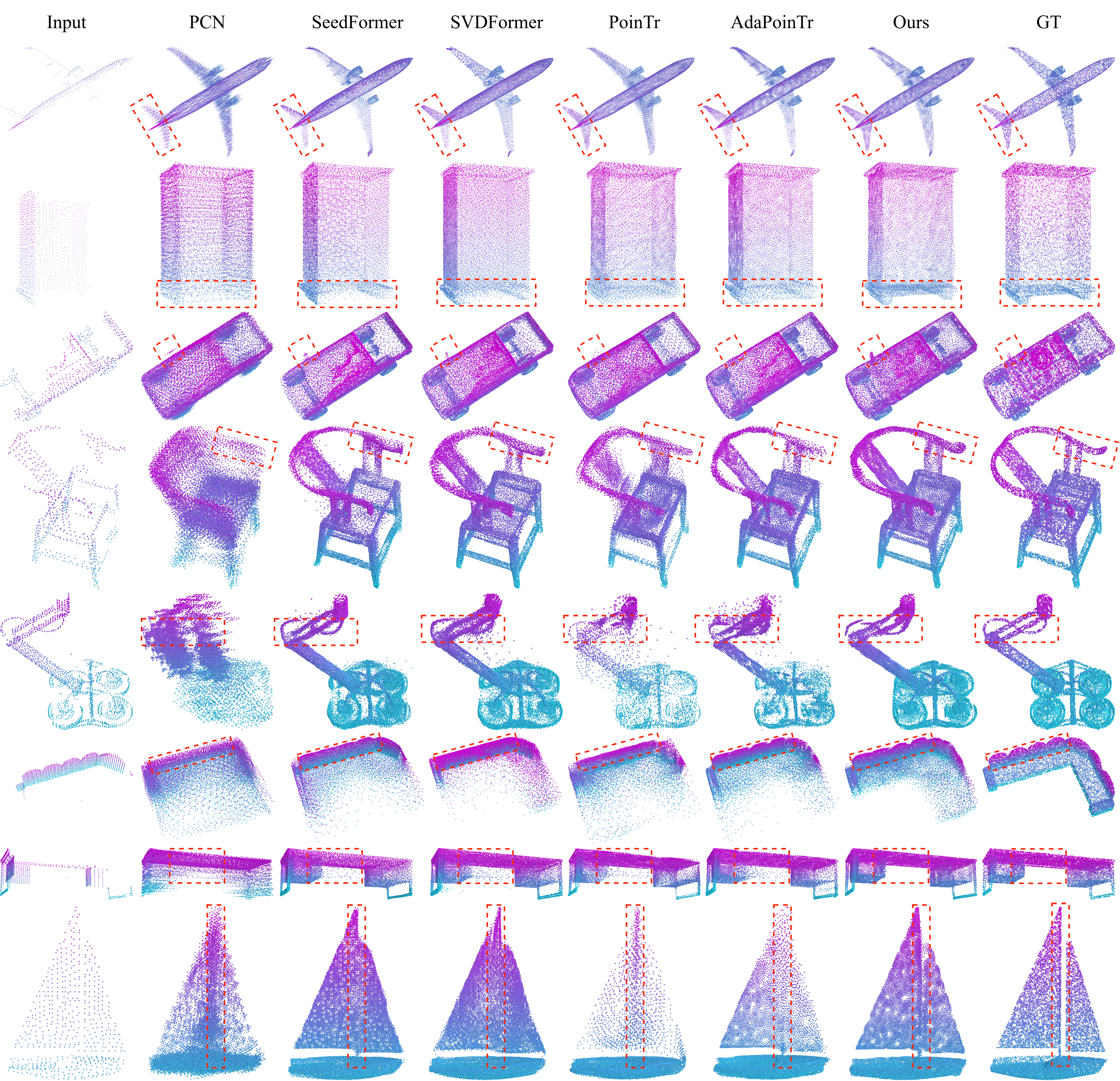}
  \caption{Qualitative comparison of PCN dataset. All methods above take the point clouds in the first column as inputs. We highlight some regions with red bounding box, which clearly show the effectiveness of our method. }
  \label{fig:pcn}
\end{figure*}

\textbf{Qualitative comparison.} Fig.~\ref{fig:pcn} visualizes the completion results of different methods on all eight categories of the PCN dataset. Compared to other methods, our approach not only achieves distributions closer to the ground truth but also demonstrates superior detail recovery capabilities (e.g., the tail wing of the airplane in the first row, the legs of the cabinet in the second row, the rearview mirror of the car in the third row, etc.).

\begin{table}[]
\caption{Results on ShapeNet55. All reported results are averaged across three difficulty levels. (CD-\(\ell_1 \times 10^3\), CD-\(\ell_2 \times 10^3\), EMD\(\times 10^3\) and F1-Score@1\%)}
\resizebox{\linewidth}{!}{%
\begin{tabular}{l|cccccc}
\hline
\textbf{Method} & \textbf{CD-\(\ell_1\) \(\downarrow\)} & \textbf{CD-\(\ell_2\)\(\downarrow\)} & \textbf{DCD\(\downarrow\)}  & \textbf{EMD\(\downarrow\)} & \textbf{F1(mCls)\(\uparrow\)} & \textbf{F1 (Ova)\(\uparrow\)} \\ \hline
FoldingNet~\cite{yang2017foldingnet}      & -               & 3.12            & 0.082          & -             & -                  & 0.082              \\
TopNet~\cite{tchapmi2019topnet}          & -               & 2.91            & 0.126          & -             & -                  & 0.126              \\
PCN~\cite{yuan2018pcn}             & -               & 2.66            & 0.618          & -             & -                  & 0.133              \\
GRNet~\cite{xie2020grnet}           & -               & 1.97            & 0.592          & -             & -                  & 0.238              \\
PoinTr~\cite{yu2021pointr}          & 13.95           & 1.05            & 0.539          & 38.86         & 0.468              & 0.499              \\
SeedFormer~\cite{zhou2022seedformer}      & 13.97           & 0.92            & 0.552          & 37.09         & 0.444              & 0.473              \\
ProxyFormer~\cite{li2023proxyformer}     & -               & 0.93            & 0.549          & -             & -                  & 0.483              \\
SVDFormer~\cite{zhu2023svdformer}       & 13.92           & 0.82            & 0.542          & 35.36         & 0.444              & 0.474              \\
AdaPoinTr~\cite{yu2023adapointr}       & 12.94           & \textbf{0.81}   & 0.635          & 27.58         & 0.402              & 0.433              \\ \hline
PPC-MT (Ours)    & \textbf{12.71}  & 0.82            & \textbf{0.523} & 27.11         & \textbf{0.509}     & \textbf{0.539}     \\ \hline
\end{tabular}}
\label{tab:results-s55}
\end{table}

\subsection{Experiments on ShapeNet-55/34}

\textbf{Dataset.} We also evaluate our model on two additional datasets proposed in PoinTr \cite{yu2021pointr}: ShapeNet-55 and ShapeNet-34. In both datasets, the input incomplete point cloud has 2,048 points, and the complete ground truth contains 8,192 points. Similar to PoinTr, during training, we randomly select a viewpoint and delete a number of points ranging from 2,048 to 6,144 (corresponding to 25\% to 75\% of the complete point cloud). The remaining points are then downsampled to 2,048 to serve as the training input. The deleted portion is downsampled to 1,536 points and considered the ground truth missing part. During testing, we select 8 fixed viewpoints and set the number of incomplete points to 2,048, 4,096, or 6,144 (representing 25\%, 50\%, or 75\% of the complete point cloud), corresponding to three difficulty levels (easy, moderate, hard). 

\textbf{Evaluation Metrics.} To maintain consistency with previous work, in addition to CD-\(\ell_1\), Earth Mover's Distance (EMD), and Density-aware Chamfer Distance (DCD), we also include CD-\(\ell_2\) as an evaluation metric. It is worth noting that the number of instances varies across different categories in the ShapeNet-55 test data and the Unseen categories of ShapeNet-34. While previous works typically reported category averages for other metrics, F-Score was commonly reported as an instance average, which could lead to confusion. Therefore, for test data with unbalanced instance counts, we report both the mean-class F1-Score (mCls) and the overall instance average F1-Score (Ova). Unless otherwise specified, all average metrics reported in this paper are category averages. 

\textbf{Quantitative comparison.} Table~\ref{tab:results-s55} and \ref{tab:results-s34} list the quantitative performance of several methods on ShapeNet-55 and ShapeNet-34, respectively. All metrics are averaged across the easy, moderate, and hard settings. Table~\ref{tab:results-s55} shows that PPC-MT performs excellently and consistently across all metrics, without significant imbalances like SVDFormer's EMD or AdaPoinTr's DCD and F1 performance. Most notably, the method shows a substantial improvement in F1 score, indicating a better balance between precision and recall in the completed point clouds and demonstrating its superior point cloud completion capability. Table~\ref{tab:results-s34} shows that on the 34 seen categories, PPC-MT achieves superior performance across all reported metrics. To evaluate generalization ability, we assessed performance on 21 unseen categories. PPC-MT continues to exhibit strong performance, achieving state-of-the-art results in CD-L1, DCD, EMD, and F1 metrics. This highlights the model's ability to effectively complete shapes from categories it did not encounter during training. These experimental results demonstrate the overall superior performance of PPC-MT, particularly its leading advantage in DCD and F1 scores, fully proving the robustness and strong generalization capacity of our method. 

\begin{table*}[ht]
\caption{Results on ShapeNet34. All reported results are averaged across three difficulty levels. (CD-\(\ell_1 \times 10^3\), CD-\(\ell_2 \times 10^3\), EMD\(\times 10^3\) and F1-Score@1\%)}
\resizebox{\linewidth}{!}{%
\begin{tabular}{l|ccccc|cccccc}
\hline
\multicolumn{1}{c|}{\multirow{2}{*}{Method}} & \multicolumn{5}{c|}{34  seen categories}                                          & \multicolumn{6}{c}{21  unseen categories}                                                         \\ \cline{2-12} 
\multicolumn{1}{c|}{}   & CD-\(\ell_1\)\(\downarrow\)   & CD-\(\ell_2\)\(\downarrow\)  & DCD\(\downarrow\)   & EMD\(\downarrow\)     & F1\(\uparrow\)            & CD-\(\ell_1\)\(\downarrow\)         & CD-\(\ell_2\)\(\downarrow\)        & DCD\(\downarrow\)           & EMD\(\downarrow\)           & F1 (mCls)\(\uparrow\)   & F1 (0va)\(\uparrow\)      \\ \hline
FoldingNet~\cite{yang2017foldingnet}                                   & -              & 2.35          & -              & -              & 0.139          & -              & 3.62          & -              & -              & -             & 0.095          \\
TopNet~\cite{tchapmi2019topnet}                                       & -              & 2.31          & -              & -              & 0.171          & -              & 3.50           & -              & -              & -             & 0.121          \\
PCN~\cite{yuan2018pcn}                                          & -              & 2.22          & 0.624          & -              & 0.154          & -              & 3.85          & 0.644          & -              & -             & 0.101          \\
GRNet~\cite{xie2020grnet}                                         & -              & 1.74          & 0.600          & -              & 0.251          & -              & 2.99          & 0.625          & -              & -             & 0.216          \\
PoinTr~\cite{yu2021pointr}                                       & 15.07          & 1.09          & 0.604          & 42.35          & 0.402          & 17.64          & 1.75          & 0.629          & 46.30           & 0.370          & 0.360           \\
SnowflakeNet~\cite{xiang2021snowflakenet}                                 & 13.80           & 0.81          & 0.594          & 27.86          & 0.414          & 14.43          & 1.00             & 0.588          & 28.03          & 0.399         & 0.378          \\
SeedFormer~\cite{zhou2022seedformer}                                   & -              & 0.83          & 0.561          & -              & 0.452          & -              & 1.34          & 0.586          & -              & -             & 0.402          \\
SVDFormer~\cite{li2023proxyformer}                                    & 13.40           & 0.74          & 0.540           & 35.35          & 0.458          & 15.53          & 1.29          & 0.555          & 38.83          & 0.424         & 0.404          \\
ProxyFormer~\cite{li2023proxyformer}                                  & -              & 0.81          & 0.556          & -              & 0.466          & -              & 1.42          & 0.583          & -              & -             & 0.415          \\
AdaPoinTr~\cite{yu2023adapointr}                                    & -              & 0.73          & -              & -              & 0.469          & -              & 1.23          & -              & -              & -             & 0.416          \\ \hline
PCD-MT (Ours)                                & \textbf{12.11} & \textbf{0.70} & \textbf{0.520} & \textbf{26.38} & \textbf{0.523} & \textbf{14.12} & \textbf{1.27} & \textbf{0.533} & \textbf{28.74} & \textbf{0.490} & \textbf{0.472} \\ \hline
\end{tabular}}
\label{tab:results-s34}
\end{table*}

\begin{figure*}
  \centering
  \includegraphics[width=\linewidth]{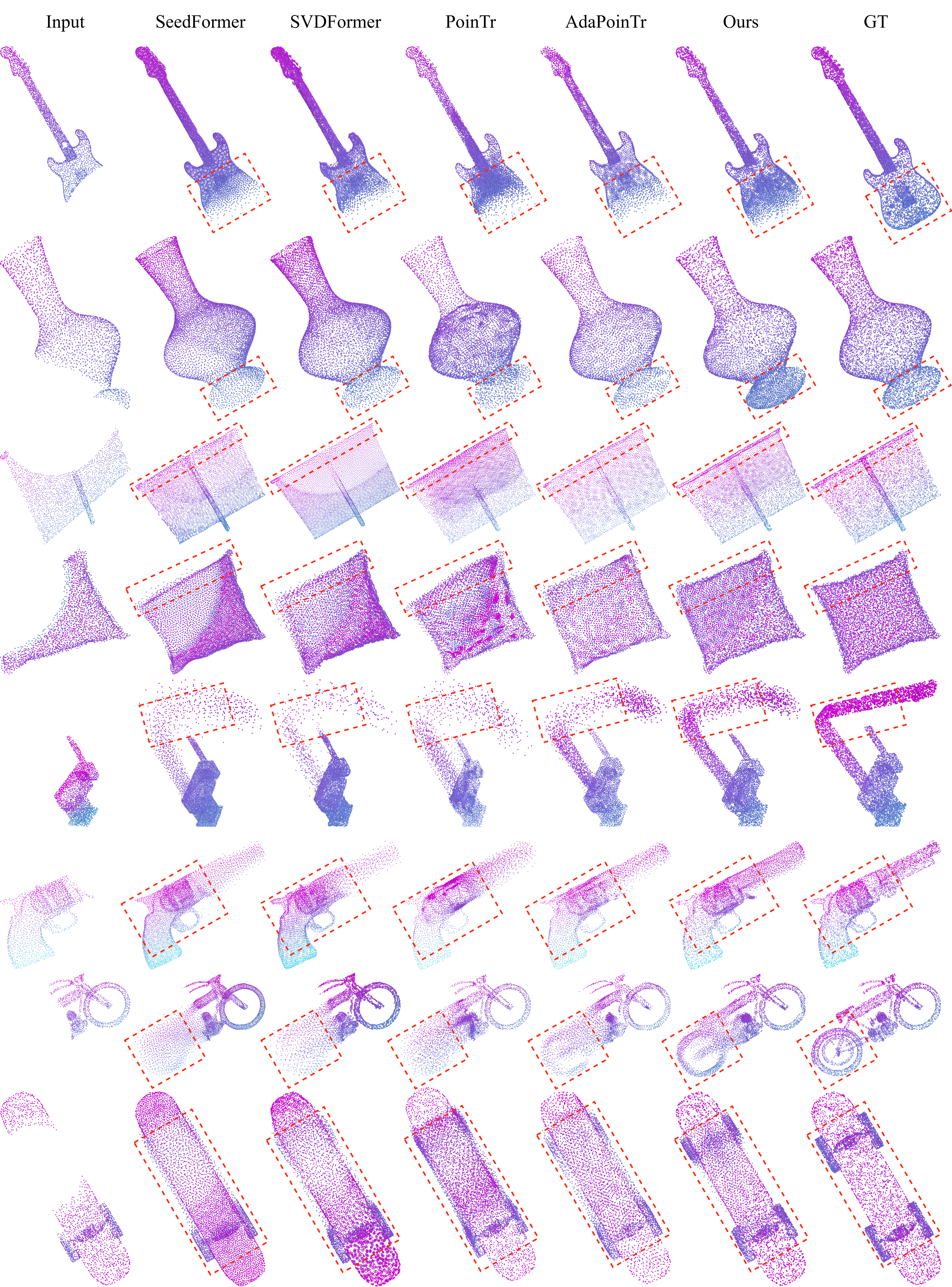}
  \caption{Qualitative comparison of ShapeNet55. All methods above take the point clouds in the first column as inputs. We highlight some regions with red bounding box, which clearly show the effectiveness of our method. }
  \label{fig:s55}
\end{figure*}

\textbf{Qualitative comparison.} Fig.~\ref{fig:s55} presents visualization results for eight categories from the ShapeNet-55 dataset that are not part of the PCN dataset. All visual results clearly indicate that our method produces results closer to the ground truth distribution with better local details. The excellent results on these less common categories further demonstrate the broad applicability of our method. 

\subsection{Experiments on KITTI}

\textbf{Dataset.} To demonstrate the generalization performance of our method in real-world scenarios, we conducted experiments on the KITTI dataset. Following previous methods \cite{xie2020grnet,yu2021pointr,yu2023adapointr}, we fine-tuned the model pre-trained on the PCN dataset using the ShapeNetCars dataset (the car subset of ShapeNet), and then evaluated its performance on the KITTI Car dataset for fair comparison.

\textbf{Evaluation Metrics.} Since the KITTI dataset lacks complete ground truth point clouds, it is challenging to evaluate the quality of completion results from a quantitative perspective. Therefore, we provide evaluation results from multiple perspectives as references. In addition to using Fidelity, Consistency, and Minimal Matching Distance (MMD) from PCN \cite{yuan2018pcn}, we also incorporate Uniformity~\cite{li2019pu,xie2020grnet} as an evaluation metric. All evaluation metrics follow the principle that lower values indicate better performance. Although GRNet mentioned that Fidelity and MMD are not well-suited for evaluating KITTI dataset results, they can still reflect the distribution characteristics of completion results to some extent, so we still provide these results. Specifically, Fidelity computes the average distance from each point in the input to its nearest neighbor in the output, measuring the degree to which the input is preserved; MMD calculates the Chamfer Distance (CD) between the output and the car point cloud in ShapeNet that is closest to the output point cloud, measuring how much the output resembles a typical car; Consistency computes the average CD between completion outputs of the same instance in consecutive frames, measuring the consistency of the network's output to input variations; Uniformity measures the overall distribution uniformity by sampling M output points (set to 1000 in this paper) and analyzing their local distributions, which can be expressed as:

\begin{equation}
\mathrm{Uniformity}(p) = \frac{1}{M}\sum_{i=1}^{M}\mathrm{U_{imbalance}}(S_i)\mathrm{U_{clutter}}(S_i)
\end{equation}

where $S_i (i = 1,2,\ldots, M)$ are point subsets obtained by ball queries with radius $\sqrt{p}$ around M seed points acquired through farthest point sampling from the output $\mathcal{P}$. $\mathrm{U_{imbalance}}$ and $\mathrm{U_{clutter}}$ represent global and local distribution uniformity, respectively.

\begin{equation}
\mathrm{U_{imbalance}}(S_i)=\frac{(\lvert{S_i}\rvert-\hat{n})^2}{\hat{n}}
\end{equation}

where $\hat{n} = p\lvert\mathcal{P}\rvert$ is the expected number of points in $S_i$.

\begin{equation}
\mathrm{U_{clutter}}(S_i)=\frac{1}{\lvert{S_i}\rvert}\sum_{j=1}^{\lvert{S_i}\rvert}\frac{(d_{i,j}-\hat{d})^2}{\hat{d}}
\end{equation}

where $d_{i,j}$ represents the distance from the $j$-th point in $S_i$ to its nearest neighbor, and $\hat{d}$ is approximately $\sqrt{\frac{2\pi p}{\lvert{S_i}\rvert\sqrt{3}}}$ when $S_i$ has a uniform distribution.

\begin{table}[t]
\caption{Results on KITTI dataset. (Fidelity \(\times 10^3\), MMD \(\times 10^3\), Consistency \(\times 10^3\))}
\resizebox{\linewidth}{!}{%
\begin{tabular}{l|cccccccc}
\hline
\multirow{2}{*}{Method} & \multicolumn{1}{l}{\multirow{2}{*}{Fidelity\(\downarrow\)}} & \multicolumn{1}{l}{\multirow{2}{*}{MMD\(\downarrow\)}} & \multicolumn{1}{l}{\multirow{2}{*}{Consistency\(\downarrow\)}} & \multicolumn{5}{c}{Uniformity\(\downarrow\) for different \(p\)}                             \\ \cline{5-9} 
                        & \multicolumn{1}{l}{}                                             & \multicolumn{1}{l}{}                      & \multicolumn{1}{l}{}                              & 0.4\%         & 0.6\%         & 0.8\%         & 1.0\%         & 1.2\%          \\ \hline
GRNet~\cite{xie2020grnet}                   & 0.816                                                            & 0.568                                     & 0.407                                             & 2.69          & 4.98          & 7.42          & 9.93          & 12.46          \\
PoinTr~\cite{yu2021pointr}                  & 0.000                                                            & 0.526                                     & 0.512                                             & 4.21          & 7.55          & 11.46         & 15.77         & 20.44          \\
AdaPoinTr~\cite{yu2023adapointr}               & 1.439                                                            & \textbf{0.366}                            & 0.569                                             & 15.17         & 21.87         & 28.56         & 35.09         & 41.62          \\ \hline
PPC-MT (Ours)                  & 1.487                                                            & 0.374                                     & \textbf{0.403}                                    & \textbf{1.46} & \textbf{3.19} & \textbf{5.45} & \textbf{8.16} & \textbf{11.24} \\ \hline
\end{tabular}}
\label{tab:results-kitti}
\end{table}
\textbf{Quantitative Comparison.} Table~\ref{tab:results-kitti} presents a quantitative comparison of the PPC-MT method against GRNet, PoinTr, and AdaPoinTr across four key evaluation metrics on the KITTI dataset. The results for GRNet and PoinTr are based on pre-trained weights provided by the original authors, while for AdaPoinTr, which lacks publicly available pre-trained models, we retrained the model following their official code to obtain the results. Regarding the Fidelity metric, PoinTr achieves a score of 0 due to its structural design that preserves all input points. The performance of PPC-MT on Fidelity and MMD indicates that its completion results tend to reconstruct input point clouds and generate results more similar to the car category in the PCN dataset. The optimal performance on Consistency demonstrates that our network is robust in maintaining output consistency. Particularly noteworthy is the outstanding performance of PPC-MT in the Uniformity metric evaluation, where it achieves the best results across all evaluated neighborhood parameters p, significantly outperforming other comparative methods. This fully demonstrates the superior capability of our method in generating uniformly distributed point clouds.

\begin{figure}[t]
  \centering
  \includegraphics[width=\linewidth]{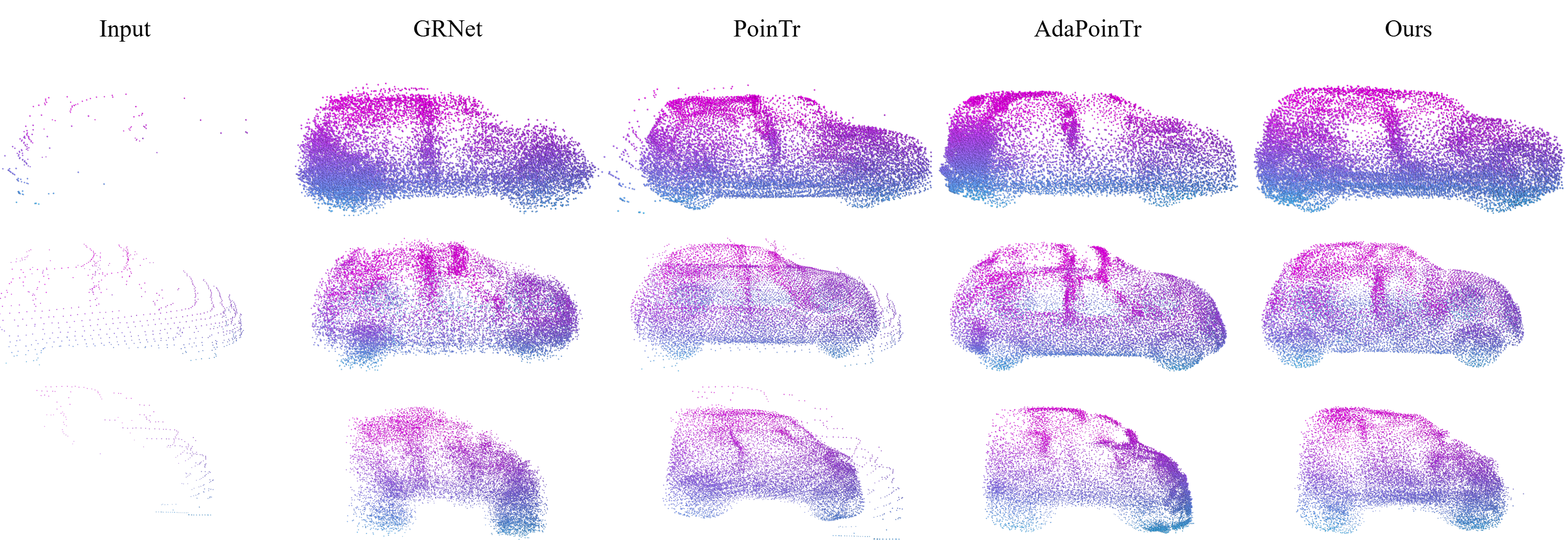}
  \caption{Qualitative comparison of KITTI. All methods above take the point clouds in the first column as inputs. }
  \label{fig:kitti}
\end{figure}

\textbf{Qualitative Comparison.} Fig.~\ref{fig:kitti} presents qualitative comparison results of various methods. From the visual effects, it can be intuitively observed that the point clouds generated by PPC-MT not only present more realistic structures in overall morphology but also exhibit smoother and clearer geometric features in local details. In contrast, other methods show certain limitations in detail fidelity and structural integrity.
        
\begin{table}[t]
\caption{Effect of different number of reconstruction heads. (CD-\(\ell_1 \times 10^3\), EMD\(\times 10^3\) and F1-Score@1\%)}
\centering
\begin{tabular}{l|cccc}
\hline
Rec. Heads & CD-\(\ell_1\)\(\downarrow\) & DCD\(\downarrow\)  & EMD\(\downarrow\)  & F1\(\uparrow\)   \\ \hline
2                   & \textbf{6.58}   & 0.502          & 18.59          & 0.857          \\
4 (ours)            & 6.60            & \textbf{0.491} & 17.43          & \textbf{0.860} \\
8                   & 6.71            & 0.494          & \textbf{17.09} & 0.855          \\ \hline
\end{tabular}
\label{tab:abl-heads}
\end{table}

\begin{table}[t]
\caption{Effect of decomposition strategy. (CD-\(\ell_1 \times 10^3\), EMD\(\times 10^3\) and F1-Score@1\%)}
\centering
\begin{tabular}{l|cccc}
\hline
Method      & CD-\(\ell_1\)\(\downarrow\)        & DCD\(\downarrow\)           & EMD\(\downarrow\)           & F-Score\(\uparrow\)       \\ \hline
Random Dec. & 6.61          & 0.495          & 17.57          & 0.858          \\
Uniform Dec. (Ours)        & \textbf{6.60} & \textbf{0.491} & \textbf{17.43} & \textbf{0.860} \\ \hline
\end{tabular}
\label{tab:abl-decomposition}
\end{table}

\subsection{Ablation Studies}
In this subsection, we conduct ablation studies on PPC-MT using the PCN dataset~\cite{yuan2018pcn} to demonstrate the effectiveness of our proposed components. All experimental settings are identical to those in Section~\ref{para-pcn} by default. 

\textbf{Number of Reconstruction Heads.} The parallel reconstruction strategy is key to our PPC-MT framework. Similar to how serial strategies determine the optimal number of completion stages, parallel strategies need to determine the optimal number of reconstruction heads. As shown in Table~\ref{tab:abl-heads}, we compare variants with different numbers of reconstruction heads. The results indicate that reconstruction performance is optimal when the number of heads is 4, achieving the best scores for DCD, EMD, and F1.

\textbf{Decomposition Strategy.} While reconstruction heads target the predicted point cloud, the decomposition strategy targets the ground truth point cloud. In our PPC-MT framework, the decomposition strategy is also crucial for the reconstruction results. To validate the effectiveness of our decomposition strategy, we compared different methods. Our framework requires the decomposed point clouds to have good global coverage. Some decomposition methods performed poorly, so we only show the comparison between random decomposition and our PCA-based sorting followed by uniform decomposition. The results in Table~\ref{tab:abl-decomposition} demonstrate the effectiveness of our proposed strategy.

\textbf{Hybrid Architecture.} To validate the effectiveness of the proposed hybrid architecture in PPC-MT, we designed a network variant where only the encoder was changed to a Transformer, keeping all other modules consistent. As shown in Table~\ref{tab:abl-architecture}, the Mamba encoder outperforms the Transformer encoder in terms of both Params and FLOPs, while also achieving better results. 
\begin{table}[t]
\caption{Effect of different architecture. (CD-\(\ell_1 \times 10^3\), EMD\(\times 10^3\) and F1-Score@1\%)}
\resizebox{\linewidth}{!}{%
\begin{tabular}{lccccccc}
\hline
Method               & Layers & Params & FLOPs  & CD-L1\(\downarrow\)        & DCD\(\downarrow\)           & EMD\(\downarrow\)          & F1\(\uparrow\)       \\ \hline
Transformer Enc.  & 6      & 62.81M & 33.48G & 6.74          & 0.509          & 19.02          & 0.846          \\
Transformer Enc.  & 12     & 75.43M & 36.71G & 6.63          & 0.494          & 17.69          & 0.857          \\ \hline
Mamba Enc. (Ours) & 12     & 55.50M & 30.68G & \textbf{6.60} & \textbf{0.491} & \textbf{17.43} & \textbf{0.860} \\ \hline
\end{tabular}}
\label{tab:abl-architecture}
\end{table}
\subsection{Complexity Analysis}
Table~\ref{tab:complexity} presents a complexity analysis, showing the number of parameters, FLOPs, and results on the PCN dataset, measured on a single NVIDIA 3090 GPU. The comparison indicates that our method achieves a favorable trade-off between cost and performance. 
\begin{table}[t]
\caption{Complexity Analysis. (CD-\(\ell_1 \times 10^3\), EMD\(\times 10^3\) and F1-Score@1\%)}
\resizebox{\linewidth}{!}{%
\begin{tabular}{l|cccccc}
\hline
Method     & Params & FLOPs  & CD-L1\(\downarrow\)         & DCD\(\downarrow\)            & EMD\(\downarrow\)            & F1\(\uparrow\)        \\ \hline
SeedFormer~\cite{zhou2022seedformer} & 3.31M  & 53.76G & 6.74          & 0.567          & 27.91          & 0.821          \\
SVDFormer~\cite{zhu2023svdformer}  & 32.63M & 39.26G & 6.54          & 0.532          & 24.58          & 0.841          \\
AdaPoinTr~\cite{yu2023adapointr}  & 32.49M & 15.08G & 6.53          & 0.536          & 24.12          & 0.845          \\ \hline
Ours       & 55.50M & 30.68G & \textbf{6.60} & \textbf{0.491} & \textbf{17.43} & \textbf{0.860} \\ \hline
\end{tabular}}
\label{tab:complexity}
\end{table}
\section{Conclusion}

In this paper, we proposed PPC-MT, a novel parallel framework specifically targeting the task of point cloud completion. By introducing a hybrid architecture based on Mamba and Transformer, along with a parallel reconstruction strategy guided by PCA-based decomposition, PPC-MT achieves state-of-the-art (SOTA) performance on multiple point cloud completion benchmarks, showing significant effectiveness, particularly in enhancing point distribution uniformity and detail recovery. This validates our method as a new solution for the point cloud completion field, balancing high efficiency with high quality.

Furthermore, we believe the core technical principles embodied in PPC-MT possess broader value for computer graphics beyond the completion task itself. The hybrid architecture provides an effective paradigm for balancing global and local feature learning. The PCA-based geometric structuring method offers new ways to handle unordered geometric data common in graphics (like point clouds and meshes). And the parallel synthesis strategy indicates a promising direction for scalably generating increasingly complex digital assets. These principles could inspire future research in various computer graphics areas, including 3D shape generation, geometric analysis, and efficient representation learning. Future work may involve exploring more advanced geometric structuring methods and extending this parallel paradigm to other 3D representations.






\newpage

\bibliographystyle{IEEEtran}
\bibliography{ref}

\begin{IEEEbiography}[{\includegraphics[width=1in,height=1.25in,clip,keepaspectratio]{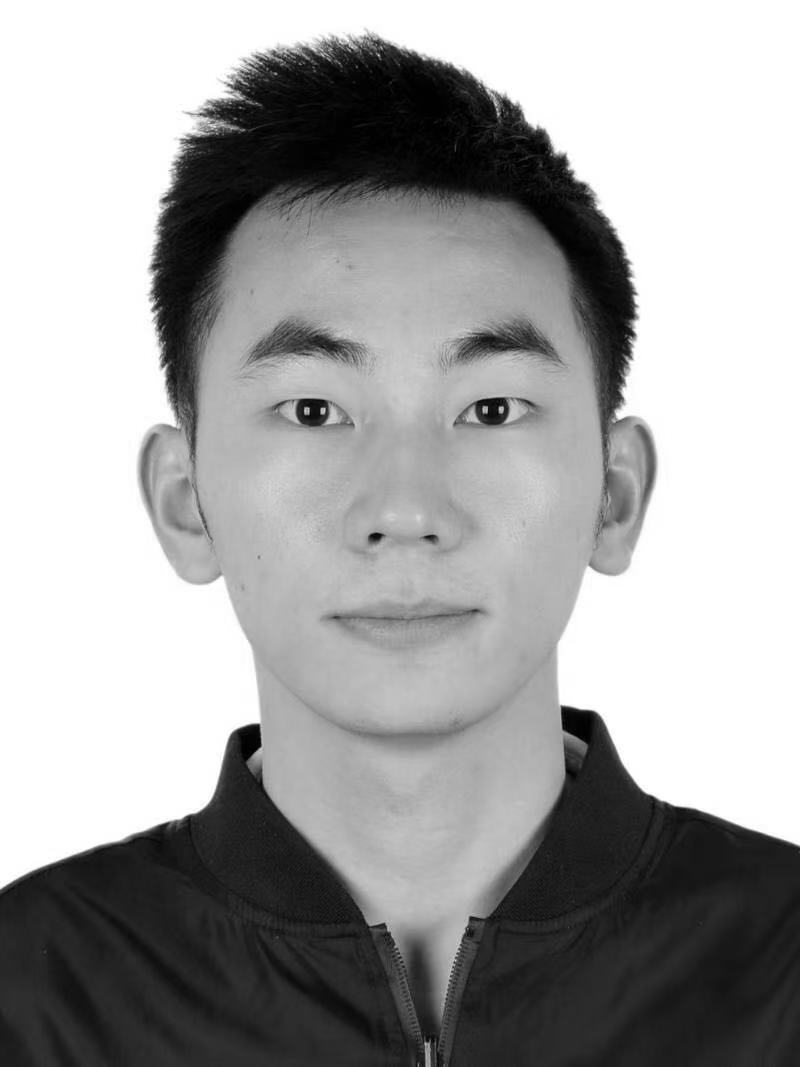}}]{Jie Li} received the B.S. degree from Hefei University of Technology, Hefei, China, in 2019. He is currently pursuing the Ph.D. degree at the College of Computer Science and Technology of Xinjiang University. His research interests include pattern recognition and 3D vision.
\end{IEEEbiography}

\begin{IEEEbiography}[{\includegraphics[width=1in,height=1.25in,clip,keepaspectratio]{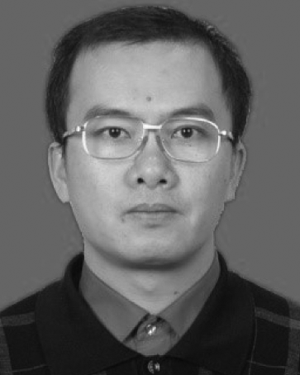}}]{Shengwei Tian} received the BS, MS, and PhD degrees from the School of Information Science and Engineering, Xinjiang University, Urumqi, China, in 1997, 2004, and 2010, respectively. Since 2002, he has been a teacher with the School of Software, Xinjiang University, where he is currently a professor. His research interests include artificial intelligence, natural language processing, and cyberspace security.
\end{IEEEbiography}

\begin{IEEEbiography}[{\includegraphics[width=1in,height=1.25in,clip,keepaspectratio]{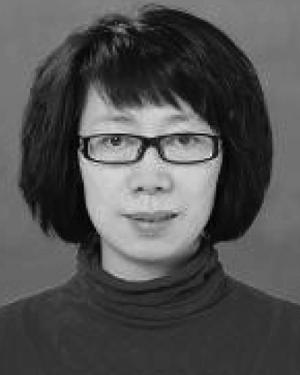}}]{Long Yu} received the B.S. and M.S. degrees from the College of Information Science and Engineering, Xinjiang University, Urumqi, China, in 1997 and 2008, respectively. Since 2002, she has been a Teacher with the College of Information Science and Engineering, Xinjiang University, where she is currently a Professor. Her research interests include artificial intelligence, data mining, natural language processing, and cyberspace security.
\end{IEEEbiography}

\begin{IEEEbiography}[{\includegraphics[width=1in,height=1.25in,clip,keepaspectratio]{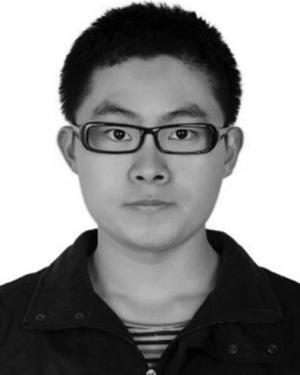}}]{Xin Ning} received his Ph.D. in 2017 from Institute of Semiconductors, Chinese Academy of Sciences. He is currently an Assistant Professor of Artificial Intelligence at Institute of Semiconductors Chinese Academy of Sciences. His research interests include deep learning machine art, pattern recognition, and image cognitive computation. He is a member of IEEE.
\end{IEEEbiography}

\end{document}